# KDH-MLTC: Knowledge Distillation for Healthcare Multi-Label Text Classification

**Hajar Sakai** and **Sarah S. Lam**
School of Systems Science and Industrial Engineering
State University of University at Binghamton
Binghamton, NY, USA
hsakai1, sarahlam@binghamton.edu

## Abstract

The increasing volume of healthcare textual data requires computationally efficient, yet highly accurate classification approaches able to handle the nuanced and complex nature of medical terminology. This research presents Knowledge Distillation for Healthcare Multi-Label Text Classification (KDH-MLTC), a framework leveraging model compression and Large Language Models (LLMs). The proposed approach addresses conventional healthcare Multi-Label Text Classification (MLTC) challenges by integrating knowledge distillation and sequential fine-tuning, subsequently optimized through Particle Swarm Optimization (PSO) for hyperparameter tuning. KDH-MLTC transfers knowledge from a more complex teacher LLM (i.e., BERT) to a lighter student LLM (i.e., DistilBERT) through sequential training adapted to MLTC that preserves the teacher's learned information while significantly reducing computational requirements. As a result, the classification is enabled to be conducted locally, making it suitable for healthcare textual data characterized by sensitivity and, therefore, ensuring HIPAA compliance. The experiments conducted on three medical literature datasets of different sizes, sampled from the Hallmark of Cancer (HoC) dataset, demonstrate that KDH-MLTC achieves superior performance compared to existing approaches, particularly for the largest dataset, reaching an F1 score of 82.70% ± 0.89%. Additionally, statistical validation and an ablation study are carried out, proving the robustness of KDH-MLTC. Furthermore, the PSO-based hyperparameter optimization process allowed the identification of optimal configurations. The proposed approach contributes to healthcare text classification research, balancing efficiency requirements in resource-constrained healthcare settings with satisfactory accuracy demands.

## 1. Introduction

The majority of healthcare facilities are known for being resource-constrained environments. Despite continuously generating data in general and unstructured data like text in particular, the lack of computational efficiency hinders the extraction of meaningful insights from it. However, the advent of Generative AI and the emergence of large language models pretrained on vast amounts of data, often referred to as Large Language Models (LLMs), are opening new possibilities and enabling the development of frameworks well-suited for healthcare settings. In parallel, the development of computational efficiency optimization methods like knowledge distillation is facilitating the use of these language models in compute-restricted scenarios. Among the key tasks that can be conducted to transform raw healthcare textual data into actionable information is Multi-Label Text Classification (MLTC).

MLTC remains a relatively challenging task where textual data like medical literature, clinical notes, patient comments, and patient call transcripts require assignment to multiple categories or concepts simultaneously. The nature of healthcare narratives, the complexity of medical terminology, and class imbalances have historically demonstrated the difficulties encountered when developing robust classification frameworks with high precision and recall across all the labels involved. LLMs allow the usage of the knowledge acquired by the model during pretraining and extend beyond domain and task-specific fine-tuning alone. Models like Bidirectional Encoder Representations from Transformers (BERT) (Devlin et al., 2018) have shown, in multiple cases, superior performance when compared to traditional Machine Learning (ML) approaches (Sung et al., 2023; Qasim et al., 2022; González-Carvajal et al., 2020). However, the computational requirements for deploying these relatively large models in resource-constrained healthcare environments present significant challenges. To address this issue, knowledge distillation (Hinton et al., 2015) has emerged as a promising technique, enabling the transfer of knowledge from a complex and higher-capacity teacher model to a simpler and more efficient student model. As a result, knowledge distillation permits not only the reduction of the model size but also retains the performance. Thereby making it particularly suitable for resource-constrained healthcare environments.

The MLTC task (i.e., Figure 1) implicitly involves dependencies across labels, requiring sequential training so the student model can learn both the teacher model's predictions and the correlations between labels. This label dependency learning is a key distinction between MLTC and single-label classification, ultimately leading to better overall (i.e., example-based) performance. Furthermore, hyperparameter optimization plays a crucial role in maximizing model performance. Trial and error selection, traditional grid search, or random search methods can be computationally expensive and inefficient and may not contribute to achieving the best performance of the approach. Particle Swarm Optimization (PSO), a population-based optimization technique inspired by the social behavior of bird flocking (Kennedy and Eberhart, 1995), offers an efficient alternative that has been successfully applied to neural network training and can either help identify the best-tuned hyperparameter values or confirm the trial and error selection.

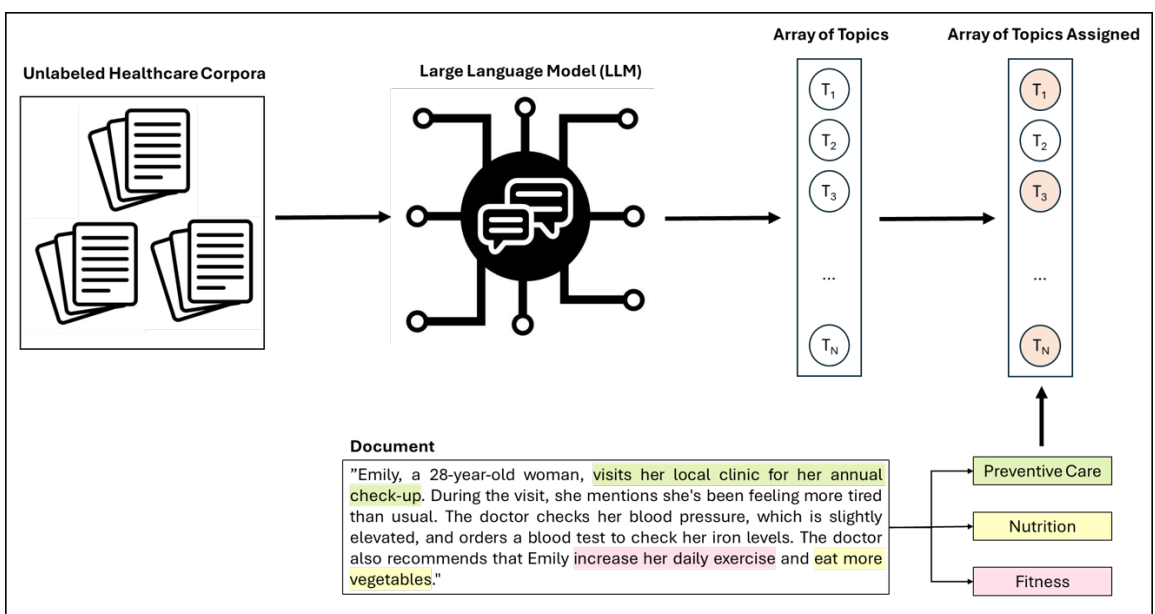

**Figure 1.** Research Problem

The literature exploration reveals a lack of sufficient research on leveraging knowledge distillation and LLMs, especially in domains where text data is nuanced, like healthcare. As a result, KDH-MLTC is introduced as an approach that not only uses BERT and one of its variants (i.e., DistilBERT) within a response-based knowledge distillation framework but also adopts sequential training to handle the multi-label nature of the task and employs PSO for hyperparameter tuning verification. Therefore, it contributes to the active research on healthcare text classification using LLMs.

This research employs the Hallmarks of Cancer (HoC) corpus (Baker et al., 2016) for approach assessment. This dataset was selected due to its prior annotation and inclusion of complex medical terminology, making it suitable for evaluation purposes. To ensure reproducibility and identify the optimal methodology for enhanced performance, evaluation was conducted across three levels. These levels comprise three distinct dataset samples (300, 500, and 1,000 text documents), resulting in varied sample sizes and topic distributions.

This research is structured as follows: The "Related Literature" section examines previous research on LLM-based classification of healthcare texts using LLMs and knowledge distillation. "Data and Sampling" describes the dataset and how samples were selected. The "Proposed Approach" section outlines the proposed KDH-MLTC methodology. The "Results and Discussion" analyzes the outcomes across binary, example-based, and level-based classification assessments, the hyperparameters tuning findings, as well as a statistical validation and an ablation study. The research concludes with "Conclusion and Future Directions," summarizing findings and suggesting areas for additional research.

## 2. Related Literature

The idea of transferring knowledge from a large and complex model (i.e., teacher model) to a smaller and simpler one (i.e., student model) by leveraging the probabilities of the teacher model (i.e., soft outputs) as targets to train the student model was first introduced by Hinton et al. (2015). It is a model compression technique where the student model is trained to mimic the teacher model behavior. It is particularly useful when the goal is to enhance the efficiency of models' deployment, especially in resource-constraint environments like healthcare settings. This framework is typically categorized into three major types: response-based, feature-based, and relation-based knowledge distillation, as shown in Figure 2.

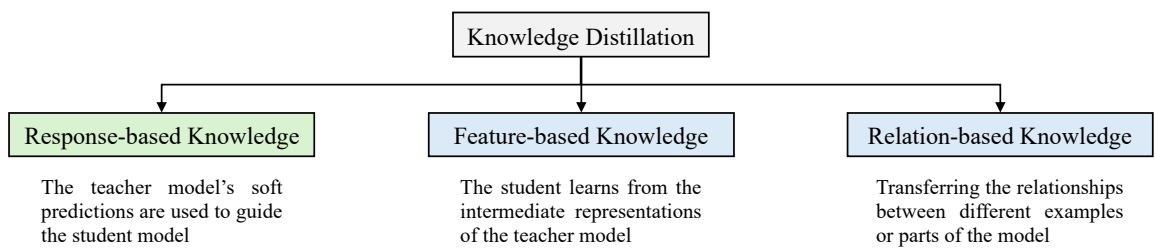

**Figure 2.** Types of Knowledge Distillation

Response-based distillation focuses on mimicking the teacher's soft output probabilities, guiding the student using softened logits to retain the teacher model's generalization capabilities (Hinton et al., 2015). This approach is characterized by its simplicity and effectiveness. Feature-based distillation, in contrast, targets the student model to replicate intermediate representations from the teacher's hidden layers (Romero et al., 2014). By aligning features or embedding spaces, this technique aims to capture deeper structural information and, therefore, enhance the student's model representations. Whereas relation-based distillation goes one step further by modeling the relationships among multiple different examples or feature representations. It mainly consists of preserving the structural topology of features, such as inter-sample distances or similarity graphs, to enable more context-aware knowledge transfer (Park et al., 2019).

In this research, response-based distillation is leveraged. It is one of the foundational methods for transferring knowledge, and its architecture is illustrated in Figure 3.

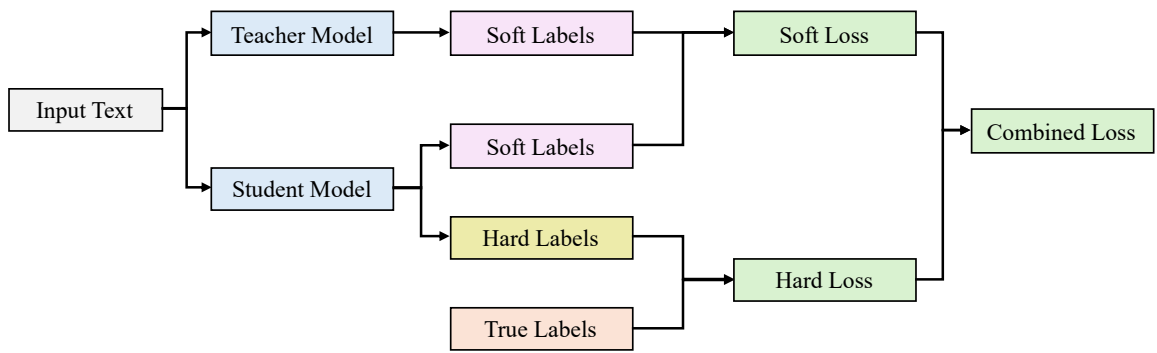

**Figure 3.** Response-based Knowledge Distillation Architecture

The student model is typically trained using a combined loss integrating two components: (1) *a soft loss*, measuring the divergence between the student's and the teacher's soft labels, often using Kullback-Leibler (KL) divergence, and

(2) *a hard loss*, comparing the student's predictions to the true labels using a commonly used classification loss function like the cross-entropy. This combination allows the student model to benefit from the richer, nuanced knowledge embedded in the teacher's softened predictions while still learning from ground truth labels (Gou et al., 2021). This strategy has been proven effective when it comes to compressing models without losing performance accuracy, as summarized in Table 1.

Table 1. Text Classification Using Knowledge Distillation in Healthcare

| Reference | Language | Data Type | Best Approach | | Results of the Best | | | | | |
| --- | --- | --- | --- | --- | --- | --- | --- | --- | --- | --- |
| | | | Teacher Model | Student Model | Dataset / Model / Label | Accuracy | F1 score | Precision | Recall | AUC score |
| Hasan et al. (2025) | English | Clinical Admission Notes | DischargeBERT + COReBERT | BERT-PKD | Diagnosis | - | - | - | - | 83.41% |
| | | | | | Procedure | - | - | - | - | 86.95% |
| | | | | | Mortality | - | - | - | - | 82.17% |
| | | | | | Length of Stay | - | - | - | - | 72.36% |
| Song et al. (2024) | English | Medical Transcriptions | Multiple RoBERTa-large | RoBERTa-large | Dataset 1 | 37.20% | 33.70% | 34.00% | 37.30% | - |
| | | | | | Dataset 2 | 37.60% | 34.00% | 35.30% | 37.70% | - |
| Ding et al. (2024) | English | Clinical Text & Structured EHR | Google T5 | Temporal Model (e.g., RNN, GRU) | Hypertension | 73.63% | 78.28% | 75.01% | 81.51% | 78.14% |
| | | | | | Heart Failure | 80.01% | 70.77% | 76.10% | 65.41% | 85.30% |
| Cho et al. (2024) | English & Korean | Clinical Notes | KM-BERT | Ko-BERT | - | 72.60% | 78.60% | 72.70% | 85.70% | 79.40% |
| Hasan et al. (2023) | English | Clinical Admission Notes | CORe + DischargeBERT | DistilBERT | Mortality | - | - | - | - | 81.34% |
| | | | | | Length of Stay | - | - | - | - | 70.56% |
| Kim and Joe (2023) | English | Medical Natural Language | BioBERT | LSTM | - | 86% | - | - | - | - |
| De Angeli et al. (2022) | English | Cancer Pathology Reports | Ensemble of 1,000 MtCNNs | Single MtCNN | Site | 96.27% | - | - | - | - |
| | | | | | Subsite | 94.82% | - | - | - | - |
| | | | | | Laterality | 96.19% | - | - | - | - |
| | | | | | Histology | 95.84% | - | - | - | - |
| | | | | | Behavior | 97.60% | - | - | - | - |
| Si et al. (2020) | English | Patient Messages | LESA-BERT | Distil-LESA-BERT-6 | - | - | 80.70% | 81.60% | 79.80% | - |

A comprehensive overview, summarized in Table 1, of the recent research studies using knowledge distillation for healthcare text classification was carried out. Hasan et al. (2025) achieved good results using DischargeBERT and COReBERT as teacher models with BERT-PKD as the student model, obtaining satisfactory AUC scores (72.36% - 86.95%) across multiple prediction tasks. Song et al. (2024) maintained poor but consistent performance in Multiple-RoBERTa-to-RoBERTa distillation for medical transcriptions across datasets, while Ding et al. (2024) successfully transferred knowledge from Google T5 to temporal models for hypertension and heart failure prediction. Cho et al. (2024) applied knowledge distillation from KM-BERT to Ko-BERT for clinical notes in a bilingual context, effectively transferring knowledge while achieving good performance. In another study, Hasan et al. (2023) distilled knowledge from CORe and DischargeBERT (i.e., teachers) to DistilBERT (i.e., student), maintaining good AUC scores for mortality and length of stay prediction. De Angeli et al. (2022) demonstrated effective knowledge transfer from an ensemble of 1000 MtCNNs to a single MtCNN for cancer pathology reports, achieving a high accuracy across all classification tasks, all being above 94%. Other notable distillation approaches include Si et al. (2020) transferring knowledge from Label Embeddings for Self-Attention in BERT (LESA-BERT) to Distil-LESA-BERT-6 for patient messages, and Kim and Joe (2023) implementing knowledge distillation from BioBERT to LSTM architectures. Based on this, several key trends emerge across diverse types of healthcare textual data. BERT variants appear to be good candidates to play both the role of the teacher and the student models. Additionally, a cross-architecture distillation (i.e., BERT-to-LSTM and ensemble-to-single model) is observed, which demonstrates flexibility. Moreover, student models achieved good performance while reducing computational demands. Furthermore, the knowledge distillation-based approaches' task sensitivity may be significant when comparing the length-of-stay prediction and other predictions for the cases of Hasan et al. (2023, 2025). The emergence of a bilingual case (Cho et al., 2024) and temporal models' integrations (Ding et al., 2024 and Kim and Joe, 2023) highlight expanding applications. However, the scoped literature reveals significant research gaps when it comes to the application of knowledge distillation for healthcare text classification, where other teacher-student combinations and integrations still require experimentation and exploration. This makes this research area a growing field aiming to balance accuracy with deployability in resource-constrained healthcare environments.

## 3. Data and Sampling

Evaluating MLTC is essential to assess how well the developed approach performs and to compare it with existing methods. However, this evaluation requires labeled data. Therefore, three sample subsets were taken from the publicly available Hallmark of Cancer (HoC) dataset (Baker et al., 2016) for this purpose.

### 3.1. Hallmark of Cancer (HoC) Dataset

Tumor cells gain various biological abilities, known as hallmarks, during cancer development. Viewing cancer through these hallmarks helps simplify the otherwise complex nature of cancer biology. Hanahan and Weinberg first presented this framework in 2000 and later revised it in 2011, and it has become highly influential in cancer research. The framework identifies 10 distinct hallmarks, which are illustrated in Figure 4.

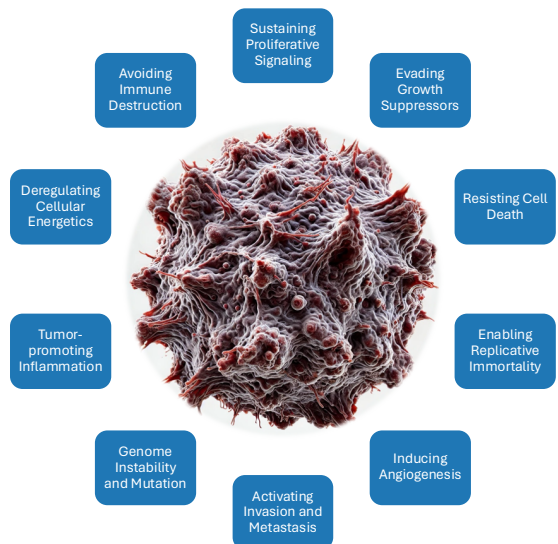

**Figure 4.** Hallmarks of Cancer (HoC)

This research examines the Hallmarks of Cancer corpus created by Baker et al. (2016), which contains 1,499 PubMed abstracts. Experts annotated these abstracts at the sentence level, focusing only on sentences presenting findings or conclusions that provide scientific evidence for any of the 10 cancer hallmarks. When a sentence offered sufficient evidence for multiple hallmarks, it received multiple labels accordingly. The researchers performed an agreement analysis on a subset of 155 abstracts to validate the annotation methodology.

### 3.2. Data Sampling

The reliability of the proposed approach was tested using three stratified samples from the HoC dataset, containing 300, 500, and 1,000 text documents, respectively. These samples were designed to maintain the original dataset's topic distribution, with the largest sample distribution represented in Figure 5. Using different sample sizes allows for the evaluation of the approach's scalability, which is especially important for LLM-based methods where performance can vary based on input size and where batch processing is often necessary due to context size limitations or in cases where fine-tuning is involved. Furthermore, this incremental assessment verifies the proposed approach's ability to generalize and its sensitivity to data volume. It also confirms whether a gain in performance is achieved when sufficient annotated data is available.

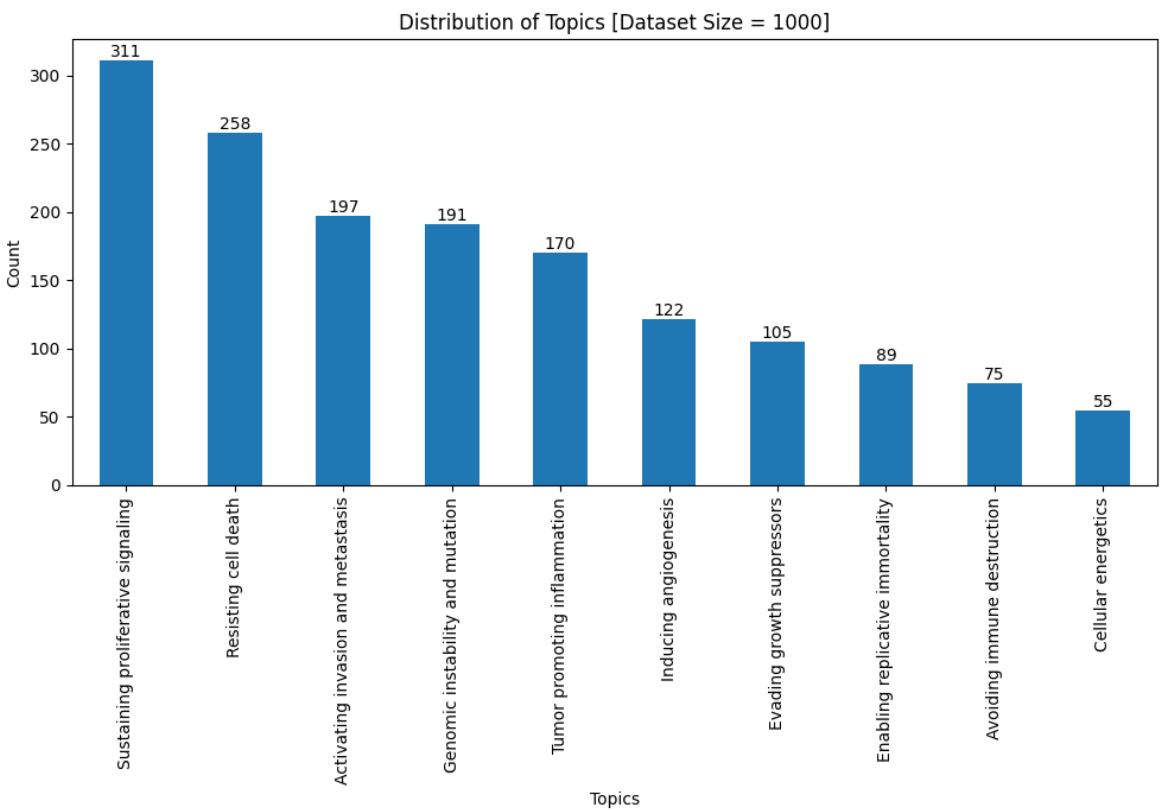

**Figure 5.** Topics Distribution

## 4. Proposed Approach

This section details the proposed approach, KDH-MLTC, using LLMs within a knowledge distillation framework for computationally efficient and accurate healthcare MLTC. The key components of this approach include the teacher-student combination selected, the sequential training process, and the use of a metaheuristic (i.e., PSO) for optimal hyperparameter verification. Figure 6 illustrates KDH-MLTC.

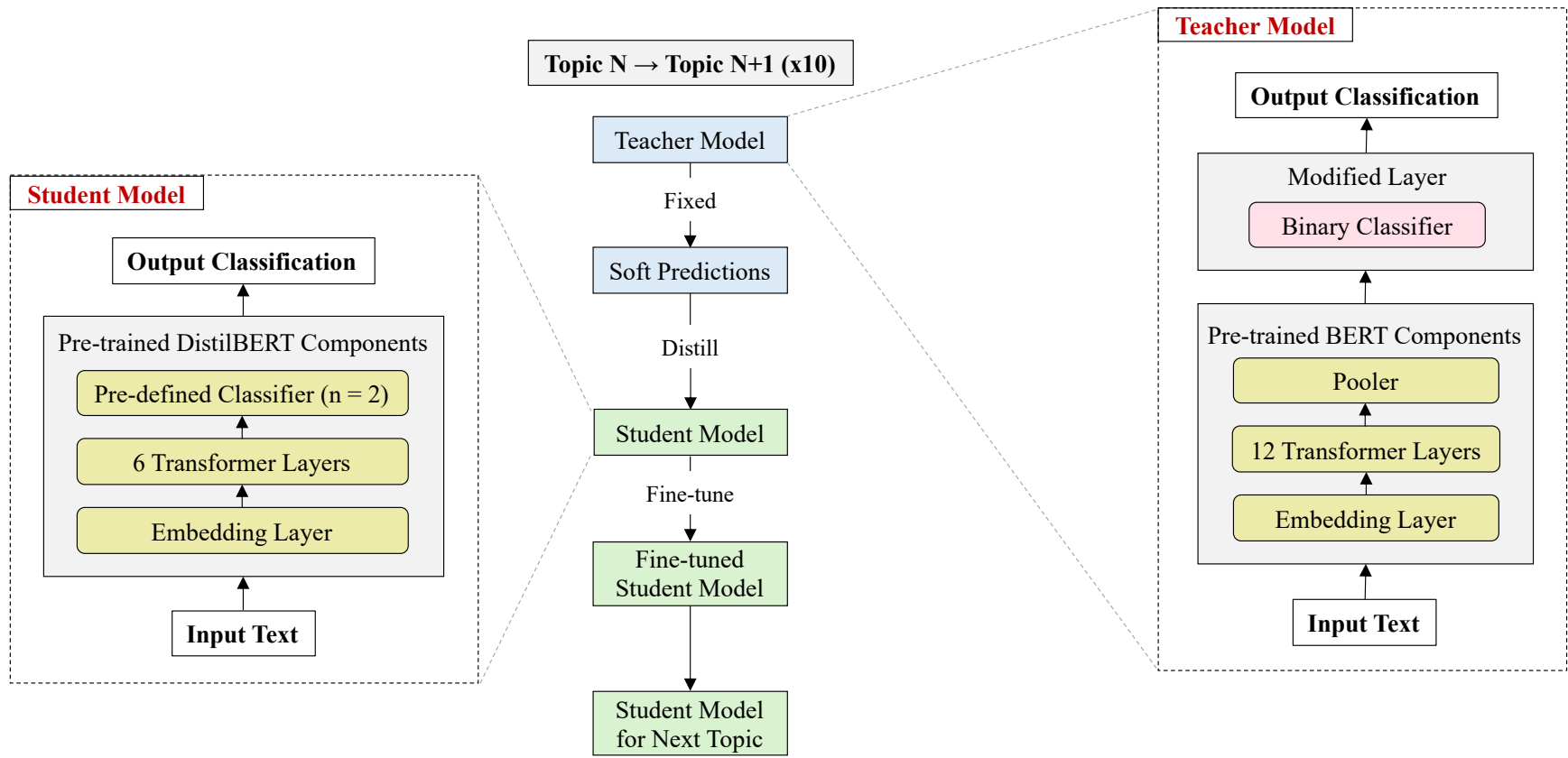

**Figure 4.** Proposed KDH-MLTC Approach

## 4.1. Teacher-Student Combination Selection

BERT-DistilBERT combination for KDH-MLTC was strategically selected to balance performance and efficiency in healthcare MLTC. The selection of BERT (Devlin et al., 2018) as the teacher model was driven by several key considerations. BERT is characterized by extensive pretraining on large text corpora, which enables it to develop a deep understanding of language semantics and contextual relationships, and this is crucial for processing complex medical terminology and concepts. Moreover, the model's architecture, containing 12 transformer layers and multi-head attention mechanisms, provides the capacity to capture complex patterns in healthcare narratives that might be missed by simpler models. Additionally, BERT's bidirectional nature allows for comprehensive text analysis that considers the full context rather than processing text in a single direction. Also, its transformer architecture enables the efficient handling of long-range dependencies in text. For the student model, DistilBERT (Sanh et al., 2019) offers multiple advantages that make it ideal for practical deployment. It maintains approximately 97% of BERT's performance capabilities while requiring only 60% of the computational resources and being 40% smaller in size. Furthermore, the architectural similarity between the teacher and student models, including the use of the same tokenizer, creates an alignment that facilitates effective knowledge transfer. Besides, with only 6 transformer layers compared to BERT's 12, DistilBERT is particularly well-suited for deployment in resource-constrained healthcare environments where computational efficiency is key.

Other potential teacher-student combinations are possible. However, each has its own drawbacks compared to the BERT-DistilBERT combination. For instance, RoBERTA-DistilRoBERTa combination would introduce unnecessary computational overhead due to RoBERTA being resource-intensive (Liu et al., 2019; Pookduang et al., 2025). The BioBERT-TinyBERT (Lee et al., 2020; Jiao et al., 2019) pairing would present challenges due to the large knowledge gap between the domain-specific teacher and the heavily compressed student, potentially compromising effective knowledge transfer. The ClinicalBERT-ALBERT (Huang et al., 2019; Lan et al., 2019) combination would suffer from architectural incompatibilities and slower student model performance that would hinder practical deployment.
re
As a result, it can be deduced that the BERT-DistilBERT combination provides a balance across a number of aspects. This includes the architectural compatibility between the selected teacher and student models, ensuring an effective knowledge transfer during the distillation process. Additionally, this pairing offers a balance between generalization capability and computational efficiency, allowing for both accurate predictions and practical deployment. BERT's strong performance as a teacher model provides a solid foundation for healthcare text classification tasks, while DistilBERT's compact architecture makes it feasible to deploy in real-world healthcare settings. As a result, the transfer of pretrained knowledge from BERT to DistilBERT enables leveraging good language understanding within practical computational constraints. These combined advantages make BERT-DistilBERT a suitable choice for the proposed KDH-MLTC.

## 4.2. Training Process

The KDH-MLTC's training process is characterized by a knowledge loss function combining soft and hard losses, as first introduced by Hinton et al. (2015). It is also defined as being sequential, taking into account the multi-label nature of the healthcare classification that has been conducted.

### 4.2.1. Knowledge Distillation Loss

The loss function used combines two components:

- **Soft Loss:** It consists of calculating the Kullback-Leibler divergence between two distributions. The temperature T controls the smoothness of the probability distributions, with high temperature values producing softer distributions, revealing more information about the relative similarities. The $T^2$ factor compensates for the softening effect.

$$\text{Loss}_{\text{soft}} = \text{KLDiv}(\log(\sigma_s), \sigma_t) \times (T^2)$$

where,
- $\sigma_s = \text{softmax}(z_s/T)$ the student's softened probability distribution
- $\sigma_t = \text{softmax}(z_t/T)$ the teacher's softened probability distribution

- **Hard Loss:** It corresponds to the standard cross-entropy loss. This is the conventional loss function used in supervised classification tasks, where the model's predictions are compared against the true labels. Unlike the soft loss comparing the softened probability distributions, the hard loss penalizes the student model based on how well it predicts the ground truth.

$$\text{Loss}_{\text{hard}} = \text{CrossEntropyLoss}(z_s, y_{\text{true}})$$

where, $z_s$ the student predictions and $y_{\text{true}}$ the true labels

- **Weighted combined loss function:** It balances the soft and hard loss components, allowing control over how much the student model should focus on aligning with the teacher's probability distribution compared to correctly classifying the ground truth labels.

$$\text{Loss}_{\text{KD}} = \alpha \, \text{Loss}_{\text{soft}} + (1 - \alpha) \, \text{Loss}_{\text{hard}}$$

### 4.2.2. Sequential Training

The training and validation process of KDH-MLTC employs a five-fold stratified cross-validation approach to ensure robust evaluation. Each of the three datasets was partitioned into five equal folds while maintaining the distribution of class labels across all of them (i.e., stratified). This helps mitigate potential bias from imbalanced data distributions and provides reliable performance metrics.

The data processing pipeline consists of three primary components. All input texts were processed as string data types and tokenized. The same tokenization was maintained for BERT and DistilBERT for compatible knowledge transfer. Topic columns were converted to binary classification targets for each of the ten distinct labels, enabling multi-label classification capabilities. Predictions were systematically gathered across all folds and topics to facilitate comprehensive performance evaluation.

The approach's architecture comprises two primary components:

- **Teacher model:**

$$h_t = \text{BERT}(X)$$

where, $h_t$ the hidden representation from 'bert-base-uncased' (hidden size, H = 768) and X the input text

Since this is the case of a classification task, a task-specific layer for the teacher model is introduced:

$$f_{\text{teacher}}(h_t) = W_t h_t + b_t, \text{ where } W_t \in \mathbb{R}^{H \times 2}$$

- **Student model:**

$$h_s = \text{DistilBERT}(X)$$

$$f_{\text{student}}(h_s) = W_s h_s + b_s$$

where, $h_s$ is derived from 'distilbert-base-uncased' and X the input text

The proposed sequential training strategy operates with a nested loop structure. The outer loop iterates through each of the five cross-validation folds, designating one fold as validation data and the remaining folds as training data. The middle loop iterates through each of the ten topics. The inner loop proceeds through all the epochs for each topic. During each epoch, the teacher model's predictions are generated. Then, the student model is trained by computing the knowledge distillation loss mentioned earlier. The obtained predictions are stored for validation and evaluation. This sequential approach ensures efficient knowledge transfer while considering the cross-label dependencies, thus improving the overall MLTC performance.

## 4.3. Hyperparameters Tuning

The results and findings reported and discussed afterward rely on a hyperparameter configuration resulting from manual trial and error. However, to ensure the optimal selection of these hyperparameters, an additional optimization layer is added by leveraging a metaheuristic for tuning.

Particle Swarm Optimization (PSO) is the metaheuristic that has been resorted to for hyperparameter tuning of KDH-MLTC. One of the major strengths that PSO offers lies in its population-based search mechanism (Kennedy & Eberhart, 1995), where particles represent hyperparameter combinations that iteratively adjust based on individual and collective experiences. This allows PSO to navigate efficiently the complex search spaces typically characterizing deep learning optimization problems. When compared to metaheuristics like Genetic Algorithm (GA), PSO demonstrates faster convergence and lower computational overhead, which are advantages particularly important given KDH-MLTC's setting. While GAs rely on stochastic operators like crossover and mutation, which can potentially lead to slower convergence and higher randomness, PSO maintains a memory of prior good solutions, enhancing efficiency. In summary, PSO provides an effective balance of exploration and exploitation, computational efficiency, and interpretability that aligns well with the specific challenges of knowledge distillation for hyperparameters optimization, as supported by multiple research studies (Lorenzo et al., 2017; Fouad et al., 2021; Tayebi and El Kafhali, 2022; Kilichev and Kim, 2023).

In the specific context of KDH-MLTC, the hyperparameters considered can be categorized into those related to the knowledge distillation loss function: the soft loss temperature (i.e. [2, 4]) and the loss function weight (i.e., [0.1, 0.9]) and those characterizing the student model, which are the learning rate (i.e., [0.0001, 0.001]), the batch size (i.e., [8, 64]), the number of epochs (i.e., [3, 5]), and the maximum sequence length for tokenization (i.e., [128, 512]).

A parallel PSO was carried out as detailed in Algorithm 1, with the fitness value being the example-based F1 score of the KDH-MLTC. Parallel PSO is likely to outperform sequential PSO by reducing computation time through the distribution of particles across multiple processes, as demonstrated in Vanneschi et al. (2011) and Zemzami et al. (2020).

The parallel PSO's parameters were selected to take the commonly used values, with the inertia weight being set to 0.7 and both the cognitive and social weights to 1.5. The number of particles (and parallel jobs) was equal to ten, as well as the number of iterations. An early stopping is also integrated with a threshold equal to 0.1% and a patience set to 1. All experiments were conducted on a MacBook Pro equipped with an Apple M2 Pro chip, featuring a 12-core CPU, a 19-core integrated GPU, and 32 GB of unified memory. The hyperparameters tuning using a five-fold iterative cross-validation and sequential training with ten labels of BERT-DistilBERT knowledge distillation took around six hours to run.

**Algorithm 1: Hyperparameter optimization using Parallel PSO**

1: PSO(h, n, w, $c_1$, $c_2$, T, N)
/* **h:** hyperparameter space, **n:** number of particles, **w:** inertia weight, **$c_1$:** cognitive coefficient, **$c_2$:** social coefficient, **T:** max iterations, **N:** early stopping rounds */
2: Initialize n particles $\{x_1, x_2, ..., x_n\} \subset h$ randomly, Initialize velocities $\{v_1, v_2, ..., v_n\}$ randomly
3: Initialize pbest_pos[i] = $x_i$ for all i ∈ [1,n], pbest_score[i] = -∞ for all i ∈ [1,n]
4: Initialize gbest_pos = null, gbest_score = -∞, no_improv_count = 0, prev_best = -∞
5: **for** t = 1 to T **do**
6:    **for** i = 1 to n **in parallel do**
7:       params = extract_hyperparams($x_i$) /* Extract hyperparameters from particle position */
8:       **for** each fold in cross_validation_folds **do** /* Evaluate using cross-validation */
9:          X_train, X_val, y_train, y_val = get_fold_data(fold)
10:         teacher_model = initialize_BERT()
11:         student_model = initialize_DistilBERT()
12:         **for** each topic in topics **do**
13:            topic_y_train = y_train[topic], topic_y_val = y_val[topic]
14:            teacher_model, student_model = train_knowledge_distillation(teacher_model,
15:               student_model, X_train, X_val, topic_y_train, topic_y_val, temperature,
16:               alpha, learning_rate, batch_size, epochs, max_length)
17:            topic_predictions = predict(student_model, X_val)
18:            store_predictions(topic, topic_predictions)
19:       **end for**
20:    **end for**
21:    score = calculate_multilabel_F1_score(all_predictions, all_actuals)
22:    **if** score > pbest_score[i] **then** /* Update personal and global best */
23:      pbest_score[i] = score
24:      pbest_pos[i] = $x_i$
25:    **end if**
26:    **if** score > gbest_score **then**
27:      gbest_score = score
28:      gbest_pos = $x_i$
29:    **end if**
30: **end for**
31: **for** i = 1 to n **do** /* Update velocities and positions */
32:    $r_1, r_2$ = random_vector(dim = dimension_of_h) /* Random values in [0,1] */
33:    $v_i = w \cdot v_i + c_1 \cdot r_1 \cdot$ (pbest_pos[i] - $x_i$) + $c_2 \cdot r_2 \cdot$ (gbest_pos - $x_i$)
34:    $x_i = x_i + v_i$
35:    $x_i$ = apply_constraints($x_i$, h)
36: **end for**
37: improvement = gbest_score - prev_best
38: **if** improvement < threshold **then** /* Check for early stopping */
39:    no_improv_count = no_improv_count + 1
40: **else**
41:    no_improv_count = 0
42: **end if**
43: prev_best = gbest_score
44: **if** no_improv_count ≥ N **then**
45:    break
46: **end if**
47: **end for**
48: **Return** gbest_pos, gbest_score

## 4.4. Performance Metrics

The performance of the new approach and existing methods is evaluated by computing the True Positive, True Negative, False Positive, and False Negative results from the confusion matrix. Since this research involves MLTC, various evaluation approaches to measure performance are used: example-based, label-based, and binary evaluations. The *example-based* metrics assess how accurately each text is classified as a whole by comparing the assigned topics with the actual topics mentioned (Sorower, 2010). In this context, the F1 score is calculated using the following formula:

$$F1 = 2 \times \frac{\text{Precision} \times \text{Recall}}{\text{Precision} + \text{Recall}} \tag{1}$$

where

$$\text{Precision} = \frac{\text{TP}}{\text{TP} + \text{FP}} \tag{2}$$

and

$$\text{Recall} = \frac{\text{TP}}{\text{TP} + \text{FN}} \tag{3}$$

The *label-based* evaluation treats each topic as an independent binary classification task, computes the relevant metrics for each, and then takes the average across all topics (Sorower, 2010). This approach calculates Micro-F1, Macro-F1, and Weighted-F1 scores according to the following methodology:

$$\text{Micro} - F1 = 2 \times \frac{\text{Micro} - \text{Precision} \times \text{Micro} - \text{Recall}}{\text{Micro} - \text{Precision} + \text{Micro} - \text{Recall}} \tag{4}$$

where

$$\text{Micro} - \text{Precision} = \frac{\sum_{j=1}^{|L|} TP_j}{\sum_{j=1}^{|L|} TP_j + \sum_{j=1}^{|L|} FP_j} \tag{5}$$

and

$$\text{Micro} - \text{Recall} = \frac{\sum_{j=1}^{|L|} TP_j}{\sum_{j=1}^{|L|} TP_j + \sum_{j=1}^{|L|} FN_j} \tag{6}$$

here, $TP_j$, $FP_j$, and $FN_j$ are true positives, false positives, and false negatives for the $j^{th}$ topic.

$$\text{Macro} - F1 = \frac{1}{T} \sum_{j=1}^{|L|} F1_j \tag{7}$$

where $F1_j$ is the F1 score of the $j^{th}$ topic.

$$\text{Weighted} - F1 = \sum_{j=1}^{|L|} w_j \times F1_j \tag{8}$$

where

$$w_j = \frac{\text{Number of instances of topic j}}{\text{Total number of texts}} \tag{9}$$

*Binary* evaluation transforms the evaluation process into a series of independent binary classifications (first-order strategy) (Zhang & Zhou, 2013), where each topic is evaluated on its own. In this approach, F1 and Area Under the Curve (AUC) metrics are used for assessment.

The various assessment techniques together offer a thorough overview of how well KDH-MLTC, as well as the other existing approaches, perform by evaluating classification accuracy at topic and text document levels.

## 5. Results and Discussion

In this section, KDH-MLTC performance is evaluated by comparing it to several other methods for MLTC. Additionally, an ablation study is conducted to demonstrate the importance of sequential training compared to binary relevance and to compare the knowledge distillation loss used with a hybrid approach that combines it with a contrastive loss. Furthermore, statistical validation is carried out and the hyperparameter tuning findings are discussed.

### 5.1. Performance Metrics Summary

Performance comparison depends on a three-level evaluation process: binary, example-based, and label-based assessments. This evaluation framework is the primary reason for sampling sets from the annotated dataset (i.e., HoC). These labeled samples serve two purposes: enabling both the training and validation of supervised methods while also allowing for the evaluation of unsupervised approaches.

#### 5.1.1. Binary Classification Evaluation

In terms of binary classification evaluation, the first contrast is with traditional ML and Pretrained Language Models (PLMs) (i.e., Table 2). The comparison between the three different approaches reveals unexpected patterns across the ten topics. The traditional TF-IDF + Lin-SVM (Classifier Chains) method consistently outperformed the more advanced transformer models, achieving F1 scores ranging from approximately 61.99% to 89.71% and AUC values between 73.84% and 91.57%. BERT, despite its popularity in conducting multiple NLP tasks, achieved poor results in this specific task application, with F1 scores varying between 0% and 26.40% and AUC score values oscillating near 50%. Meanwhile, BART demonstrated somewhat better results than BERT, yet significantly underperformed when compared to traditional ML. BART's F1 scores are characterized by a wider range, varying from 12.36% to 67.66%, with AUC scores between 46.01% and 77.24%. In terms of topics, the TF-IDF + Lin-SVM (Classifier Chains) method performed best on "Resisting cell death", while BERT achieved its highest F1 score on "Sustaining proliferative signaling" and BART on "Inducing angiogenesis". Additionally, all three models struggled with common topics: "Avoiding immune destruction" and "Cellular energetics". As shown in Figure 5, both these topics are characterized by being minority labels, which explains why the traditional ML and BERT performed poorly, given that the reported scores are the results of a five-fold iterative stratified cross-validation. It is worth noting that HuggingFace was used for both PLMs (i.e., bert-base-uncased and bart-large-mnli). It is worth noting that all these evaluations are done using the largest set (i.e., 1,000).

Table 2. Traditional Machine Learning and Pretrained Language Models Results

| Supervised Learning | TF-IDF + Lin-SVM (Classifier Chains) | | BERT | | BART | |
|---|---|---|---|---|---|---|
| Topics/Metrics | F1 | AUC | F1 | AUC | F1 | AUC |
| Sustaining proliferative signaling | 78.98% | 84.12% | 26.40% | 47.88% | 27.98% | 53.17% |
| Resisting cell death | 89.71% | 91.57% | 21.73% | 48.00% | 32.48% | 46.01% |
| Activating invasion and metastasis | 87.36% | 89.86% | 15.29% | 49.31% | 58.33% | 73.04% |
| Genomic instability and mutation | 84.30% | 88.88% | 18.64% | 50.60% | 49.82% | 67.09% |
| Tumor promoting inflammation | 78.60% | 82.76% | 13.19% | 49.62% | 30.30% | 58.51% |
| Inducing angiogenesis | 82.46% | 85.54% | 14.68% | 52.00% | 67.66% | 77.24% |
| Evading growth suppressors | 61.99% | 74.51% | 11.32% | 51.77% | 18.48% | 53.04% |
| Enabling replicative immortality | 77.03% | 81.91% | 7.09% | 50.23% | 22.43% | 56.41% |
| Avoiding immune destruction | 63.16% | 73.84% | 4.92% | 49.62% | 16.93% | 57.17% |
| Cellular energetics | 72.09% | 78.18% | 0.00% | 48.31% | 12.36% | 54.60% |

Investigating the In-Context Learning (ICL) results with GPT-4o across different numbers of shots (i.e., 1, 3, and 5), a significant jump in performance can be observed when classifying the ten labels representing cancer hallmarks. The metrics generally surpassed BERT and BART, and in some cases even the TF-IDF + Lin-SVM (Classifier Chains) approach. Additionally, consistency across the different shot settings can be remarked, indicating that GPT-4o can effectively carry out the MLTC task with minimal examples. It particularly achieved very good results for "Inducing angiogenesis" and "Activating invasion and metastasis". Comparing these results to the previous supervised learning approaches, GPT-4o significantly outperformed both BERT and BART across all topics. It also exceeded the traditional ML results in several topics. However, no consistent performance improvement can be observed when increasing the number of examples from 1-shot to 5-shot. It is worth noting that, so far, all the reported F1 and AUC scores refer to the largest set (i.e., 1,000).

Table 3. In-Context Learning Results

| In-Context Learning | GPT-4o 1-shot | | GPT-4o 3-shot | | GPT-4o 5-shot | |
|---|---|---|---|---|---|---|
| Topics/Metrics | F1 | AUC | F1 | AUC | F1 | AUC |
| **Sustaining proliferative signaling** | 53.39% | 67.71% | 63.47% | 73.40% | 63.72% | 73.60% |
| **Resisting cell death** | 86.42% | 89.58% | 86.71% | 89.99% | 85.48% | 89.71% |
| **Activating invasion and metastasis** | 92.43% | 94.53% | 90.81% | 93.38% | 91.62% | 94.12% |
| **Genomic instability and mutation** | 86.49% | 90.71% | 86.79% | 91.29% | 87.47% | 92.02% |
| **Tumor promoting inflammation** | 65.64% | 74.85% | 64.86% | 74.58% | 68.89% | 77.24% |
| **Inducing angiogenesis** | 94.65% | 97.12% | 94.65% | 97.12% | 94.17% | 96.68% |
| **Evading growth suppressors** | 58.12% | 78.97% | 51.09% | 70.75% | 50.27% | 70.74% |
| **Enabling replicative immortality** | 73.10% | 79.61% | 74.83% | 81.03% | 58.46% | 71.37% |
| **Avoiding immune destruction** | 76.13% | 88.20% | 77.71% | 90.02% | 80.52% | 90.91% |
| **Cellular energetics** | 82.09% | 98.73% | 81.48% | 98.67% | 79.71% | 98.51% |

The comparison between GPT-4o 0-shot and KDH-MLTC across different dataset sizes (i.e., 300, 500, and 1,000 samples) reveals interesting performance patterns in hallmarks of cancer MLTC this time. For the smaller dataset, GPT-4o 0-shot clearly outperforms the proposed approach across nearly all topics. This suggests that GPT-4o can effectively leverage its pretrained knowledge to classify the topic without necessarily requiring an ICL, outperforming the KDH-MLTC, which appears to struggle with the limited data size. As the dataset size increases to 500 samples, a shift in the performance is observed. While GPT-4o maintains its dominance in most topics, KDH-MLTC starts showing competitive or superior performance in certain topics, particularly "Sustaining proliferative signaling," "Resisting cell death," and "Enabling replicative immortality." This indicates that KDH-MLTC benefits significantly from increased training data. This increasing trend becomes even more pronounced with 1,000 text documents, where KDH-MLTC outperforms GPT-4o in 9 out of 10 topics based on F1 scores. KDH-MLTC shows remarkable improvements in "Enabling replicative immortality" and "Cellular energetics", demonstrating that with sufficient training data, KDH-MLTC can effectively capture the nuances of the hallmarks of cancer. GPT-4o still maintains strong performance but does not scale with increased data size, which was expected for a 0-shot approach that doesn't require additional training. These findings highlight an important trade-off in the approach selection. Table 7 summarizes the results across the three datasets by reporting the corresponding means and standard deviations. The overall results are highly impacted by the metrics differences resulting from the dataset sizes. While GPT-4o 0-shot offers performance that remains relatively stable across dataset sizes, KDH-MLTC demonstrates superior performance when sufficient training data is available.

Table 4. GPT-4o 0-shot vs KDH-MLTC Results (Set size = 300)

| Approach (Set size = 300) | GPT-4o 0-shot | | KDH-MLTC | |
|---|---|---|---|---|
| Topics/Metrics | F1 | AUC | F1 | AUC |
| **Sustaining proliferative signaling** | **55.81%** | **69.32%** | 51.85% | 67.10% |
| **Resisting cell death** | **92.31%** | **94.81%** | 86.79% | 89.68% |
| **Activating invasion and metastasis** | **90.57%** | **93.63%** | 71.26% | 78.30% |
| **Genomic instability and mutation** | **81.43%** | **85.94%** | 79.71% | 84.62% |
| **Tumor promoting inflammation** | **63.92%** | **73.63%** | 40.91% | 62.78% |
| **Inducing angiogenesis** | **94.12%** | **98.01%** | 83.12% | 88.25% |
| **Evading growth suppressors** | **53.33%** | **75.63%** | 19.35% | 55.36% |
| **Enabling replicative immortality** | **69.77%** | **79.45%** | 26.67% | 57.82% |
| **Avoiding immune destruction** | **63.16%** | **82.24%** | 57.89% | 70.97% |
| **Cellular energetics** | **66.67%** | **87.89%** | 13.33% | 53.57% |

Table 5. GPT-4o 0-shot vs KDH-MLTC Results (Set size = 500)

| Approach (Set size = 500) | GPT-4o 0-shot | | KDH-MLTC | |
|---|---|---|---|---|
| Topics/Metrics | F1 | AUC | F1 | AUC |
| **Sustaining proliferative signaling** | 58.96% | 70.62% | **66.43%** | **75.02%** |
| **Resisting cell death** | 88.06% | 91.84% | **90.37%** | **93.61%** |
| **Activating invasion and metastasis** | **95.88%** | **98.21%** | 73.68% | 83.76% |
| **Genomic instability and mutation** | **83.65%** | 88.37% | 83.02% | **88.45%** |
| **Tumor promoting inflammation** | **64.86%** | **74.95%** | 58.28% | 71.94% |
| **Inducing angiogenesis** | **96.30%** | **98.78%** | 93.85% | 95.87% |
| **Evading growth suppressors** | **51.28%** | **75.89%** | 31.58% | 60.44% |
| **Enabling replicative immortality** | 71.64% | 79.05% | **89.74%** | **92.47%** |
| **Avoiding immune destruction** | **76.54%** | **92.67%** | 71.88% | 82.21% |
| **Cellular energetics** | **81.36%** | **98.84%** | 75.00% | 81.15% |

Table 6. GPT-4o 0-shot vs KDH-MLTC Results (Set size = 1,000)

| Approach (Set size = 1,000) | GPT-4o 0-shot | | KDH-MLTC | |
|---|---|---|---|---|
| Topics/Metrics | F1 | AUC | F1 | AUC |
| **Sustaining proliferative signaling** | 54.91% | 68.76% | **74.19%** | **81.24%** |
| **Resisting cell death** | 85.04% | 87.82% | **93.62%** | **95.75%** |
| **Activating invasion and metastasis** | 84.27% | 87.51% | **89.25%** | **91.57%** |
| **Genomic instability and mutation** | 85.09% | 89.80% | **90.22%** | **92.78%** |
| **Tumor promoting inflammation** | 65.10% | 74.29% | **83.77%** | **87.40%** |
| **Inducing angiogenesis** | **94.56%** | **96.08%** | 91.36% | 94.92% |
| **Evading growth suppressors** | 57.39% | **78.13%** | 59.26% | 75.10% |
| **Enabling replicative immortality** | 70.00% | 77.42% | **96.13%** | **98.60%** |
| **Avoiding immune destruction** | 78.48% | **90.20%** | 82.52% | 88.85% |
| **Cellular energetics** | 81.82% | **97.87%** | 91.43% | 93.53% |

Table 7. GPT-4o 0-Shot vs KDH-MLTC Results (Overall)

| Approach (Overall) | GPT-4o 0-Shot | | KDH-MLTC | |
|---|---|---|---|---|
| Topics/Metrics | F1 (Mean ± SD) | AUC (Mean ± SD) | F1 (Mean ± SD) | AUC (Mean ± SD) |
| **Sustaining proliferative signaling** | 56.56% ± 0.021 | 69.57% ± 0.010 | **64.16%** ± 0.113 | **74.45%** ± 0.071 |
| **Resisting cell death** | 88.47% ± 0.036 | 91.49% ± 0.035 | **90.26%** ± 0.034 | **93.02%** ± 0.031 |
| **Activating invasion and metastasis** | **90.24%** ± 0.058 | **93.12%** ± 0.054 | 78.07% ± 0.098 | 84.54% ± 0.067 |
| **Genomic instability and mutation** | 83.39% ± 0.018 | 88.04% ± 0.020 | **84.32%** ± 0.054 | **88.62%** ± 0.041 |
| **Tumor promoting inflammation** | **64.63%** ± 0.006 | **74.29%** ± 0.007 | 60.98% ± 0.216 | 74.04% ± 0.124 |
| **Inducing angiogenesis** | **94.99%** ± 0.012 | **97.62%** ± 0.014 | 89.44% ± 0.056 | 93.01% ± 0.042 |
| **Evading growth suppressors** | **54.00%** ± 0.031 | **76.55%** ± 0.014 | 36.73% ± 0.204 | 63.63% ± 0.103 |
| **Enabling replicative immortality** | 70.47% ± 0.010 | 78.64% ± 0.011 | **70.85%** ± 0.384 | **82.96%** ± 0.220 |
| **Avoiding immune destruction** | **72.73%** ± 0.083 | **88.37%** ± 0.054 | 70.76% ± 0.123 | 80.68% ± 0.090 |
| **Cellular energetics** | **76.61%** ± 0.086 | **94.87%** ± 0.061 | 59.92% ± 0.412 | 76.08% ± 0.205 |

### 5.1.2. Example-based and Label-based Evaluation

The consolidated comparison summarized in Table 8 expands on the binary evaluation of all the approaches previously discussed, focusing on example-based and label-based assessments. It provides a complementary perspective on approach performance across different dimensions of the MLTC. Examining the overall performance across approaches, KDH-MLTC, in the case of the largest dataset, emerges as outperforming across all metrics, achieving the highest scores in example-based F1 (i.e., 82.42%) and all label-based measures (i.e., Micro F1: 84.96%, Macro F1: 85.17%, Weighted F1: 84.82%). Therefore, this shows that with sufficient training data, KDH-MLTC excels at correctly classifying individual documents as well as accurately identifying each of the specific ten labels. The traditional ML approach demonstrates remarkably good performance, positioning it as the second-best approach overall. However, the PLMs considered here—BERT and BART—did perform poorly across all evaluation metrics. On the one hand, BERT, particularly, shows a relatively significant gap between example-based F1 (i.e., 14.69%) and label-based Micro F1 (i.e., 17.38%), suggesting it struggled especially with correctly classifying complete text documents that often contain multiple labels. On the other hand, BART demonstrates an even more pronounced difference between example-based F1 (i.e., 28.87%) and weighted F1 (i.e., 36.48%). Among the ICL approaches, GPT-4o with varying numbers of examples (i.e., 1, 3, and 5) shows consistent performance in the 73% - 77% range across both example-based and label-based metrics. The 3-shot and 5-shot settings slightly outperform the 1-shot approach, but the close metrics' values suggest that GPT-4o reaches a stagnation in performance with minimal examples. These values are not very different from those obtained when evaluating GPT-4o 0-shot. Furthermore, the progression of KDH-MLTC performance across the different dataset sizes showcases not only an overall improvement when more annotated data is available but also specific patterns in this amelioration. With the smallest dataset, the large gap between the example-based F1 (i.e., 53.91%) and label-based Micro F1 (i.e., 64.91%) can mean that the knowledge distillation sequential fine-tuning approach initially struggles with capturing all the labels to be assigned to each text document. As the dataset size increases to 1,000 observations, the different metrics' values become closer, showing that the KDH-MLTC was able to learn the different aspects of the MLTC.

Table 8. Example-based and Label-based Results

| Category | Approach/Metrics | Example-based F1 | Label-based Micro F1 | Label-based Macro F1 | Label-based Weighted F1 |
|---|---|---|---|---|---|
| Traditional ML | TF-IDF + Lin-SVM (Classifier Chains) | 76.63% | 80.84% | 77.57% | 80.43% |
| PLM | BERT | 14.69% | 17.38% | 13.33% | 16.92% |
| | BART | 28.87% | 32.29% | 33.68% | 36.48% |
| ICL | GPT-4o 1-Shot | 74.77% | 74.99% | 74.56% | 73.87% |
| | GPT-4o 3-Shot | 76.29% | 78.06% | 77.24% | 77.24% |
| | GPT-4o 5-Shot | 76.42% | 77.67% | 76.03% | 76.77% |
| LLM | GPT-4o 0-Shot | 75.25% | 75.56% | 75.67% | 74.46% |
| KDH-MLTC (Size = 300) | | 53.91% | 64.91% | 53.09% | 61.39% |
| KDH-MLTC (Size = 500) | | 69.09% | 75.50% | 73.38% | 74.30% |
| KDH-MLTC (Size = 1,000) | | **82.42%** | **84.96%** | **85.17%** | **84.82%** |
| KDH-MLTC (Overall) | | 68.47% | 75.13% | 70.55% | 73.50% |

## 5.2. Hyperparameters Tuning Results

Figure 7 shows the example-based F1 scores as PSO's number of iterations increases in the hyperparameter tuning process conducted for configuration verification. The graph shows a significant initial jump in F1 score between iterations 1 and 2, followed by a consistent but more gradual improvement through subsequent iterations, finally reaching 83.41% at iteration 7 and then 8, which led to an early stopping. This pattern was expected since it represents the typical convergence behavior of PSO, where more significant gains are achieved early as particles explore the search space before carrying on with refinements as they converge toward an optimal solution.

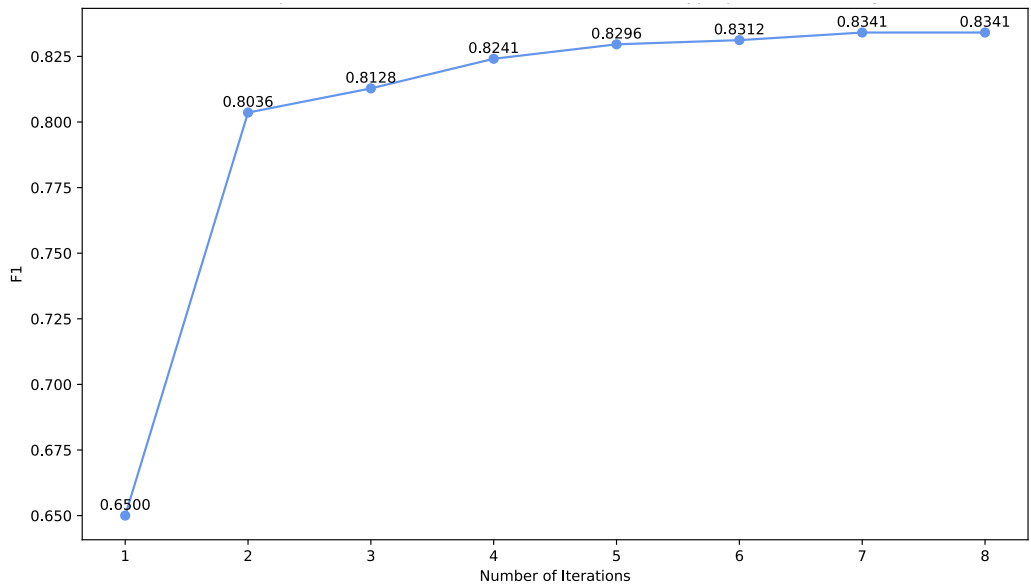

Figure 5. Example-based F1 Score Across PSO Iterations

Table 9 provides a comparison of the hyperparameter values selected through traditional trial and error and used so far for all the previous assessments versus those identified through PSO optimization. PSO identified a higher optimal temperature, suggesting that a more gradual softening would improve the performance of this MLTC. Additionally, PSO reduced the knowledge distillation loss function weight from α = 0.5 to α = 0.1, indicating that placing less emphasis on the soft loss led to better results. Also, the metaheuristic halved the batch size from 16 to 8 and extended the maximum sequence length from 128 to 512 tokens. These optimized hyperparameters resulted in a modest improvement of around 1% that can be considered a good enhancement in the context of an already high-performing approach. Moreover, the application of Parallel PSO for hyperparameter optimization demonstrates the value of systematic exploration of the parameter space beyond manual tuning. These findings further strengthen the deduction stating that KDH-MLTC is outperforming the other approaches in the case of the hallmarks of cancer MLTC.

**Table 9.** Hyperparameters Details

| Hyperparameters | Trial and Error Selection | PSO Selection |
|---|---|---|
| **Soft Loss Temperature** | T = 2 | T = 2.79 |
| **Loss Function Weight** | α = 0.5 | α = 0.1 |
| **Learning Rate** | $lr = 2 \times 10^{-5}$ | $lr = 10^{-5}$ |
| **Batch Size** | batch_size = 16 | batch_size = 8 |
| **Max Sequence Length for Tokenization** | max_length = 128 | max_length = 512 |
| **Epochs** | epochs = 5 | epochs = 5 |
| **F1 Score** | 82.42% | 83.41% |

### 4.6.3. Statistical Validation

This section provides valuable insights into the reliability and consistency of KDH-MLTC when compared to the other MLTC approaches across five replications. This statistical validation offers an additional understanding of each model's performance beyond the metrics of a single run. As shown in Table 10, summarizing the resulting descriptive statistics, KDH-MLTC emerges as the consistently best-performing approach, achieving the highest mean F1 score (i.e., 82.70%) across the replications. It is also characterized by a relatively tight confidence interval (i.e., 81.59% - 83.80%) and a modest standard deviation (i.e., 0.89%). The boxplot, illustrated in Figure 8, shows that KDH-MLTC not only achieves the highest median performance but also maintains the highest minimum F1 score, indicating its robust reliability even in its worst-case scenario. The traditional ML approach shows remarkable consistency, reaching the lowest standard deviation (i.e., 0.34%) among all methods. Among the different GPT-4o settings, the 3-shot demonstrates the best balance of performance and consistency, achieving an F1 score mean of 74.57% and a standard deviation of 0.97%, while the 5-shot reaches a similar mean performance of 74.15% with higher variability (i.e., 1.30%), and the 1-shot one shows the widest spread among all ICL variants and 0-shot setting, while this latter maintains competitive performance. Furthermore, BERT, on its own, represents a poor performance with considerable variability when conducting the hallmarks of cancer MLTC. Meanwhile, BART shows no variation across replications, indicating a deterministic behavior. The boxplot in Figure 8 highlights these patterns, emphasizing a clear separation between the high-performing approaches (i.e., KDH-MLTC, Traditional ML, and GPT-4o) and the low-performing ones (i.e., PLMs). This statistical validation reinforces previous conclusions while providing additional information. KDH-MLTC delivers the highest overall performance, and the TF-IDF + Lin-SVM (Classifier Chains) offers the best consistency. 0-shot and ICL approaches present a middle ground with good performance and reasonable stability.

Table 10. F1 score Replications Descriptive Statistics

| Approach | Mean | Std Dev | Min | Max | 95% CI Lower | 95% CI Upper |
|---|---|---|---|---|---|---|
| **Traditional ML** | 76.58% | **0.34%** | 75.98% | 76.83% | 76.15% | 77.00% |
| **BERT** | 15.04% | 1.42% | 13.03% | 16.95% | 13.27% | 16.81% |
| **BART** | 28.87% | 0.00% | 28.87% | 28.87% | - | - |
| **GPT-4o 1-shot** | 71.50% | 1.89% | 70.37% | 74.77% | 69.15% | 73.85% |
| **GPT-4o 3-shot** | 74.57% | 0.97% | 73.99% | 76.29% | 73.36% | 75.77% |
| **GPT-4o 5-shot** | 74.15% | 1.30% | 73.31% | 76.42% | 72.53% | 75.76% |
| **GPT-4o 0-shot** | 72.80% | 1.44% | 71.46% | 75.25% | 71.02% | 74.59% |
| **KDH-MLTC** | **82.70%** | 0.89% | **81.87%** | **84.21%** | **81.59%** | **83.80%** |

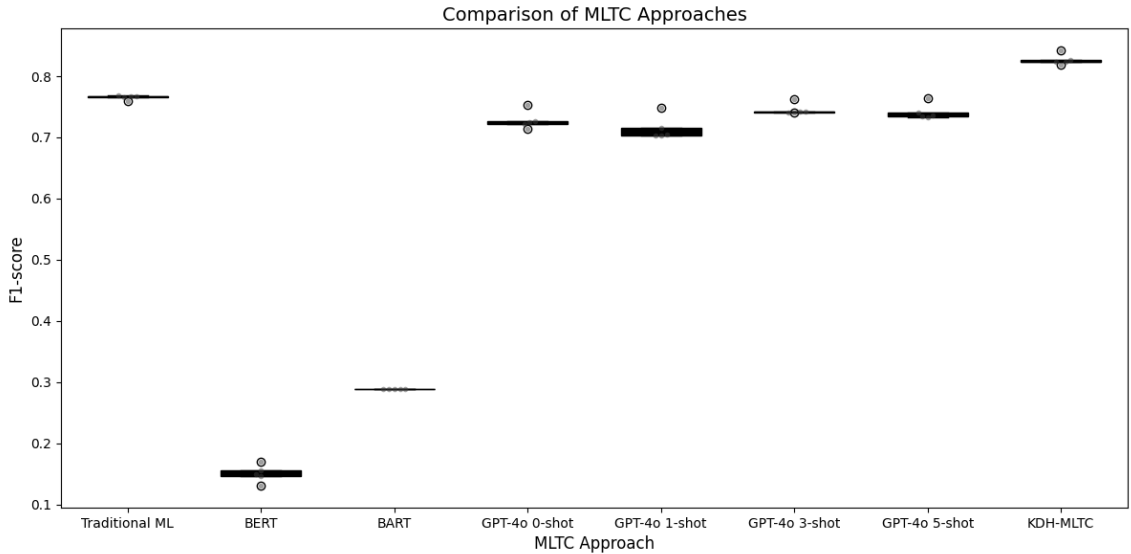

**Figure 6.** MLTC Approaches Comparison: Boxplots

The t-test results provide additional statistical evidence that KDH-MLTC significantly outperforms all other approaches in the case of this MLTC. All tests led to p-values of 0.00, indicating strong statistical significance across all comparisons. The mean differences and t-statistics reveal the magnitude of KDH-MLTC's advantage over each of the other methods. The largest performance gaps exist between KDH-MLTC and the PLMs. Additionally, KDH-MLTC maintains a statistically significant superiority even compared to other strong-performing approaches. The mean difference compared to the traditional ML approach is 0.06, supported by a significant t-statistic of -14.39. Also, KDH-MLTC outperforms all GPT-4o variants with mean differences varying between -0.11 and -0.08 and corresponding t-statistics between -11.97 and -13.84. These findings provide additional evidence that KDH-MLTC outperforms the other approaches in this research case study. It combines high performance metrics with good consistency across replications as well as statistical significance.

Table 11. t-tests Results

| Approach | Mean Difference | t-statistic | p-value | Significance |
|---|---|---|---|---|
| BERT | -0.68 | -90.19 | 0.00 | True |
| BART | -0.54 | -135.41 | 0.00 | True |
| GPT-4o 1-shot | -0.11 | -11.97 | 0.00 | True |
| GPT-4o 0-shot | -0.10 | -13.07 | 0.00 | True |
| GPT-4o 5-shot | -0.09 | -12.15 | 0.00 | True |
| GPT-4o 3-shot | -0.08 | -13.84 | 0.00 | True |
| Traditional ML | -0.06 | -14.39 | 0.00 | True |

Table 12, ANOVA's results, provide more statistical evidence of significant performance differences among the MLTC approaches contrasted, with a very high F-statistic (i.e., 2269.94), a p-value of 0.00, and a perfect effect size ($\eta^2 = 1.00$). This latter, in particular, indicates that 100% of the observed performance variation can be attributed to differences between approaches rather than random variation within approaches. As a result, this validates the previous findings that KDH-MLTC significantly outperforms all the other approaches, followed by the traditional ML approach, then GPT-4o variants, with BERT and BART substantially underperforming.

Table 12. ANOVA Results

| Metric | Value |
|---|---|
| F-statistic | 2269.94 |
| p-value | 0.00 |
| $\eta^2$ | 1.00 |

### 4.6.4. Ablation Study

Tables 13-16 summarize the ablation study results in the cases where the Knowledge Distillation (KD) loss is replaced with a combined loss integrating the knowledge distillation one with a contrastive one, and the sequential training is replaced with a binary relevance to handle the MLTC. The 'KD & Contrastive Loss' is a weighted combination of the loss used in KD-MLTC and a loss aligning the student and teacher's last hidden state embeddings using normalized similarity. The 'Binary Relevance' is breaking the MLTC into ten independent binary classification tasks, ignoring the relationships between labels. This was experimented with to assess the trade-off between the performance that the sequential training contributes to achieving versus its more complex computation efficiency.

Table 13. Performance Comparison of Approaches (F1 scores) (Set size = 300)

| Approach (Size = 300) | F1 | Micro F1 | Macro F1 | Weighted F1 |
|---|---|---|---|---|
| KD Loss + Binary Relevance | 48.48% | 64.48% | 55.60% | 62.00% |
| KD & Contrastive Loss + Binary Relevance | 33.56% | 50.13% | 33.87% | 42.50% |
| KD & Contrastive Loss + Sequential Training | 53.91% | 55.83% | 40.67% | 51.09% |
| KDH-MLTC | 53.91% | 64.91% | 53.09% | 61.39% |

Table 14. Performance Comparison of Approaches (F1 scores) (Set size = 500)

| Approach (Size = 500) | F1 | Micro F1 | Macro F1 | Weighted F1 |
|---|---|---|---|---|
| KD Loss + Binary Relevance | 57.76% | 76.22% | 73.27% | 75.39% |
| KD & Contrastive Loss + Binary Relevance | 44.52% | 61.24% | 46.65% | 56.50% |
| KD & Contrastive Loss + Sequential Training | 62.14% | 69.00% | 61.55% | 67.29% |
| KDH-MLTC | 69.09% | 75.50% | 73.38% | 74.30% |

Table 15. Performance Comparison of Approaches (F1 scores) (Set size = 1,000)

| Approach (Size = 1,000) | F1 | Micro F1 | Macro F1 | Weighted F1 |
|---|---|---|---|---|
| KD Loss + Binary Relevance | 66.99% | 84.96% | 83.92% | 84.66% |
| KD & Contrastive Loss + Binary Relevance | 63.71% | 81.96% | 78.04% | 80.37% |
| KD & Contrastive Loss + Sequential Training | 81.94% | 83.75% | 83.26% | 83.82% |
| KDH-MLTC | 82.42% | 84.96% | 85.17% | 84.82% |

Table 16. Performance Comparison of Approaches (F1 scores) (Overall)

| Approach (Overall) | F1 (Mean ± SD) | Micro F1 (Mean ± SD) | Macro F1 (Mean ± SD) | Weighted F1 (Mean ± SD) |
|---|---|---|---|---|
| KD Loss + Binary Relevance | 57.74% ± 0.093 | **75.22%** ± 0.103 | **70.93%** ± 0.143 | **74.02%** ± 0.114 |
| KD & Contrastive Loss + Binary Relevance | 47.26% ± 0.153 | 64.45% ± 0.162 | 52.85% ± 0.227 | 59.79% ± 0.191 |
| KD & Contrastive Loss + Sequential Training | 62.89% ± 0.187 | 69.52% ± 0.140 | 61.83% ± 0.213 | 67.40% ± 0.164 |
| KDH-MLTC | **68.47%** ± 0.143 | 75.13% ± 0.100 | 70.55% ± 0.162 | 73.50% ± 0.117 |

These results provide important insights into how different training approaches and loss functions affect performance across various dataset sizes. The experiments that were conducted systematically isolate the contributions of the key components: KD loss and sequential training. KDH-MLTC consistently achieves the best example-based F1 scores across dataset sizes, demonstrating its effectiveness. Furthermore, sequential training consistently outperforms binary relevance when using the same loss function, proving its relevance when dealing with text classification with a multi-label nature. The combination of knowledge distillation and contrastive loss proves beneficial only when paired with sequential training. As dataset size increases from 300 to 1,000 samples, all approaches show improved performance. Notably, for the largest dataset, the performance gaps, especially in terms of label-based assessment, become narrower. These findings demonstrate that the KDH-MLTC model provides the most balanced and generally highest performance across all datasets in terms of both example-based and label-based metrics.

## 6. Conclusion and Future Directions

This research developed an approach to efficiency optimization when using LLMs for healthcare MLTC through model compression, with a specific focus on knowledge distillation. KDH-MLTC strategically leverages sequential fine-tuning for both the teacher model (i.e., BERT) and the student model (i.e., DistilBERT), establishing a framework that balances computational efficiency with classification performance. The experimental results conclusively demonstrate that knowledge distillation combined with sequential training achieves optimal performance when sufficient annotated data is available. The proposed KDH-MLTC framework has proven particularly effective, consistently outperforming traditional machine learning algorithms, PLMs, and various single-model methodologies across multiple evaluation metrics. Additionally, PSO successfully helps verify the selected hyperparameters' values.

The concluded insights contribute to the broader research revolving around more efficient deep learning approaches for text classification tasks in resource-constrained environments like healthcare. Building upon the promising results demonstrated by KDH-MLTC, several directions for future research aiming to further enhance its performance, efficiency, and applicability can be explored. This includes the integration of knowledge distillation with federated learning frameworks. This hybridization would address privacy concerns handled by decentralized training techniques while leveraging the efficiency benefits of model compression. Since the goal is to conduct a text classification in healthcare settings, implementing KDH-MLTC within federated learning architecture could potentially enable collaborative model training across distributed datasets without requiring direct data sharing between participants.

# References

[1] Devlin, J., Chang, M. W., Lee, K., & Toutanova, K. (2018). BERT: Pre-training of deep bidirectional transformers for language understanding. arXiv preprint arXiv:1810.04805.
[2] Sung, Y. W., Park, D. S., & Kim, C. G. (2023). A Study of BERT-Based Classification Performance of Text-Based Health Counseling Data. Cmes-Computer Modeling in Engineering & Sciences, 135(1).
[3] Qasim, R., Bangyal, W. H., Alqarni, M. A., & Ali Almazroi, A. (2022). A fine-tuned BERT-based transfer learning approach for text classification. Journal of healthcare engineering, 2022(1), 3498123.
[4] González-Carvajal, S., & Garrido-Merchán, E. C. (2020). Comparing BERT against traditional machine learning text classification. arXiv preprint arXiv:2005.13012.
[5] Hinton, G., Vinyals, O., & Dean, J. (2015). Distilling the knowledge in a neural network. arXiv preprint arXiv:1503.02531.
[6] Kennedy, J., & Eberhart, R. (1995, November). Particle swarm optimization. In Proceedings of ICNN'95-international conference on neural networks (Vol. 4, pp. 1942-1948). IEEE.
[7] Sanh, V., Debut, L., Chaumond, J., & Wolf, T. (2019). DistilBERT, a distilled version of BERT: smaller, faster, cheaper and lighter. arXiv preprint arXiv:1910.01108.
[8] Baker, S., Silins, I., Guo, Y., Ali, I., Högberg, J., Stenius, U., & Korhonen, A. (2016). Automatic semantic classification of scientific literature according to the hallmarks of cancer. Bioinformatics, 32(3), 432-440.
[9] Romero, A., Ballas, N., Kahou, S. E., Chassang, A., Gatta, C., & Bengio, Y. (2014). Fitnets: Hints for thin deep nets. arXiv preprint arXiv:1412.6550.
[10] Park, W., Kim, D., Lu, Y., & Cho, M. (2019). Relational knowledge distillation. In Proceedings of the IEEE/CVF conference on computer vision and pattern recognition (pp. 3967-3976).
[11] Gou, J., Yu, B., Maybank, S. J., & Tao, D. (2021). Knowledge distillation: A survey. International Journal of Computer Vision, 129(6), 1789-1819.
[12] Hasan, M. J., Rahman, F., & Mohammed, N. (2025). OptimCLM: Optimizing clinical language models for predicting patient outcomes via knowledge distillation, pruning and quantization. International Journal of Medical Informatics, 195, 105764.
[13] Song, Y., Zhang, J., Tian, Z., Yang, Y., Huang, M., & Li, D. (2024). LLM-based privacy data augmentation guided by knowledge distillation with a distribution tutor for medical text classification. arXiv preprint arXiv:2402.16515.
[14] Ding, S., Ye, J., Hu, X., & Zou, N. (2024). Distilling the knowledge from large-language model for health event prediction. Scientific Reports, 14(1), 30675.
[15] Cho, S., Jeon, J., Lee, D., Lee, C., & Kim, J. (2024). DSG-KD: Knowledge distillation from domain-specific to general language models. IEEE Access.
[16] Hasan, M. J., Rahman, F., & Mohammed, N. (2023). Distilling the Knowledge of Clinical Outcome Predictions in Large Language Models for Resource Constrained Healthcare Systems. Available at SSRN 4591013.
[17] Kim, H., & Joe, I. (2022). Effective SNOMED-CT Concept Classification from Natural Language using Knowledge Distillation. In Proceedings of the Computational Methods in Systems and Software (pp. 54-64). Cham: Springer International Publishing.
[18] De Angeli, K., Gao, S., Blanchard, A., Durbin, E. B., Wu, X. C., Stroup, A., ... & Yoon, H. J. (2022). Using ensembles and distillation to optimize the deployment of deep learning models for the classification of electronic cancer pathology reports. JAMIA open, 5(3), ooac075.
[19] Si, S., Wang, R., Wosik, J., Zhang, H., Dov, D., Wang, G., & Carin, L. (2020, September). Students need more attention: Bert-based attention model for small data with application to automatic patient message triage. In Machine Learning for Healthcare Conference (pp. 436-456). PMLR.
[20] Sorower, M. S. (2010). A literature survey on algorithms for multi-label learning. Oregon State University, Corvallis, 18(1), 25.
[21] Zhang, M. L., & Zhou, Z. H. (2013). A review on multi-label learning algorithms. IEEE transactions on knowledge and data engineering, 26(8), 1819-1837.
[22] Liu, Y., Ott, M., Goyal, N., Du, J., Joshi, M., Chen, D., Levy, O., Lewis, M., Zettlemoyer, L., & Stoyanov, V. (2019). Roberta: A robustly optimized BERT pretraining approach. arXiv preprint arXiv:1907.11692.
[23] Pookduang, P., Klangbunrueang, R., Chansanam, W., & Lunrasri, T. (2025). Advancing Sentiment Analysis: Evaluating RoBERTa against Traditional and Deep Learning Models. Engineering, Technology & Applied Science Research, 15(1), 20167-20174.
[24] Lee, J., Yoon, W., Kim, S., Kim, D., Kim, S., So, C. H., & Kang, J. (2020). BioBERT: a pre-trained biomedical language representation model for biomedical text mining. Bioinformatics, 36(4), 1234-1240.

[25] Jiao, X., Yin, Y., Shang, L., Jiang, X., Chen, X., Li, L., Wang, F., & Liu, Q. (2019). TinyBERT: Distilling BERT for natural language understanding. arXiv preprint arXiv:1909.10351.
[26] Huang, K., Altosaar, J., & Ranganath, R. (2019). ClinicalBERT: Modeling clinical notes and predicting hospital readmission. arXiv preprint arXiv:1904.05342.
[27] Lan, Z., Chen, M., Goodman, S., Gimpel, K., Sharma, P., & Soricut, R. (2019). ALBERT: A lite BERT for self-supervised learning of language representations. arXiv preprint arXiv:1909.11942.
[28] Kennedy, J., & Eberhart, R. (1995, November). Particle swarm optimization. In Proceedings of ICNN'95-international conference on neural networks (Vol. 4, pp. 1942-1948). IEEE.
[29] Lorenzo, P. R., Nalepa, J., Kawulok, M., Ramos, L. S., & Pastor, J. R. (2017, July). Particle swarm optimization for hyper-parameter selection in deep neural networks. In Proceedings of the genetic and evolutionary computation conference (pp. 481-488).
[30] Tayebi, M., & El Kafhali, S. (2022). Deep neural networks hyperparameter optimization using particle swarm optimization for detecting frauds transactions. In Advances on Smart and Soft Computing: Proceedings of ICACIn 2021 (pp. 507-516). Springer Singapore.
[31] Fouad, Z., Alfonse, M., Roushdy, M., & Salem, A. B. M. (2021). Hyper-parameter optimization of convolutional neural network based on particle swarm optimization algorithm. Bulletin of Electrical Engineering and Informatics, 10(6), 3377-3384.
[32] Kilichev, D., & Kim, W. (2023). Hyperparameter optimization for 1D-CNN-based network intrusion detection using GA and PSO. Mathematics, 11(17), 3724.
[33] Vanneschi, L., Codecasa, D., & Mauri, G. (2011). A comparative study of four parallel and distributed PSO methods. New generation computing, 29, 129-161.
[34] Zemzami, M., El Hami, N., Itmi, M., & Hmina, N. (2020). A comparative study of three new parallel models based on the PSO algorithm. International Journal for Simulation and Multidisciplinary Design Optimization, 11, 5.